\documentclass[journal,twoside,web]{ieeecolor}
\usepackage{generic}
\usepackage{cite}
\usepackage{amsmath,amssymb,amsfonts}
\usepackage{algorithmic}
\usepackage{graphicx}
\usepackage{textcomp}

\usepackage{url}
\usepackage{cite}
\usepackage{bm}
\usepackage{hyperref}
\usepackage{amsmath}
\usepackage{amsfonts}
\usepackage{amssymb}
\usepackage{booktabs,multirow}
\usepackage{array}
\usepackage{subcaption}

\def\BibTeX{{\rm B\kern-.05em{\sc i\kern-.025em b}\kern-.08em
    T\kern-.1667em\lower.7ex\hbox{E}\kern-.125emX}}
\begin{document}
\title{Attention-Guided Discriminative Region Localization and Label Distribution Learning for Bone Age Assessment}
\author{Chao Chen, Zhihong Chen, Xinyu Jin, Lanjuan Li, William Speier, and Corey W. Arnold
\thanks{This work was supported by China Scholarship Council}
\thanks{W. Speier and C.W. Arnold are with the Computational Diagnostics Lab, the Department of Bioengineering, the Department of Radiology, and the Department of Pathology at the University of California Los Angeles, 924 Westwood Blvd, Suite 420, CA 90024 USA (e-mail: speier@ucla.edu, cwarnold@ucla.edu). }
\thanks{Chao Chen, Zhihong Chen, and Xinyu Jin are with Information Science and Electrical Engineering Department, Zhejiang Univeristy, 310027 Hangzhou, China. (e-mail: chench@zju.edu.cn, Zhihongchen@zju.edu.cn, jinxy@zju.edu.cn).}
\thanks{Lanjuan Li with State Key Laboratory for Diagnosis and Treatment of Infectious Disease, The First Affiliated Hospital, College of Medicine, Zhejiang University, 310003 Hangzhou, China(e-mail: ljli@zju.edu.cn).}}

\maketitle

\begin{abstract}
Bone age assessment (BAA) is clinically important as it can be used to diagnose endocrine and metabolic disorders during child development. Existing deep learning based methods for classifying bone age use the global image as input, or exploit local information by annotating extra bounding boxes or key points. However, training with the global image underutilizes discriminative local information, while providing extra annotations is expensive and subjective. In this paper, we propose an attention-guided approach to automatically localize the discriminative regions for BAA without any extra annotations. Specifically, we first train a classification model to learn the attention maps of the discriminative regions, finding the hand region, the most discriminative region (the carpal bones), and the next most discriminative region (the metacarpal bones). Guided by those attention maps, we then crop the informative local regions from the original image and aggregate different regions for BAA.  Instead of taking BAA as a general regression task, which is suboptimal due to the label ambiguity problem in the age label space, we propose using joint age distribution learning and expectation regression, which makes use of the ordinal relationship among hand images with different individual ages and leads to more robust age estimation. Extensive experiments are conducted on the RSNA pediatric bone age data set. Using no training annotations, our method achieves competitive results compared with existing state-of-the-art semi-automatic deep learning-based methods that require manual annotation. Code is available at \url{https://github.com/chenchao666/Bone-Age-Assessment}.
\end{abstract}

\begin{IEEEkeywords}
Bone Age Assessment, Hand Radiograph, Attention Map, Discriminative Region Localization, Label Distribution Learning.
\end{IEEEkeywords}

\section{Introduction}
Bone age assessment (BAA) from hand radiograph images is a common technique for investigating endocrinology and growth disorders \cite{poznanski1978carpal}, or for determining the final adult height of children \cite{carty2002assessment}. In clinical practice, BAA is usually performed by examining the ossification patterns in a radiograph of the non-dominant hand, and then comparing the estimated bone age with the chronological age. A discrepancy between the two values indicates abnormalities \cite{spampinato2017deep}. The most widely used manual BAA methods are Greulich-Pyle (GP) \cite{greulich1959} and Tanner-Whitehouse (TW) \cite{carty2002assessment}. In the GP method, bone age is estimated by comparing the whole hand radiograph with a reference atlas of representative ages, while the TW method examines 20 specific regions of interest (RoIs) and assigns scores based on a detailed local structural analysis. The TW method is more reliable, but time consuming, while the GP method is relatively quick and easy to use. In both manual solutions, reliable and accurate bone age estimation is limited by the subjective influence of a trained radiologist.

\begin{figure}[!t]
\begin{center}
\includegraphics[width=1.0\linewidth]{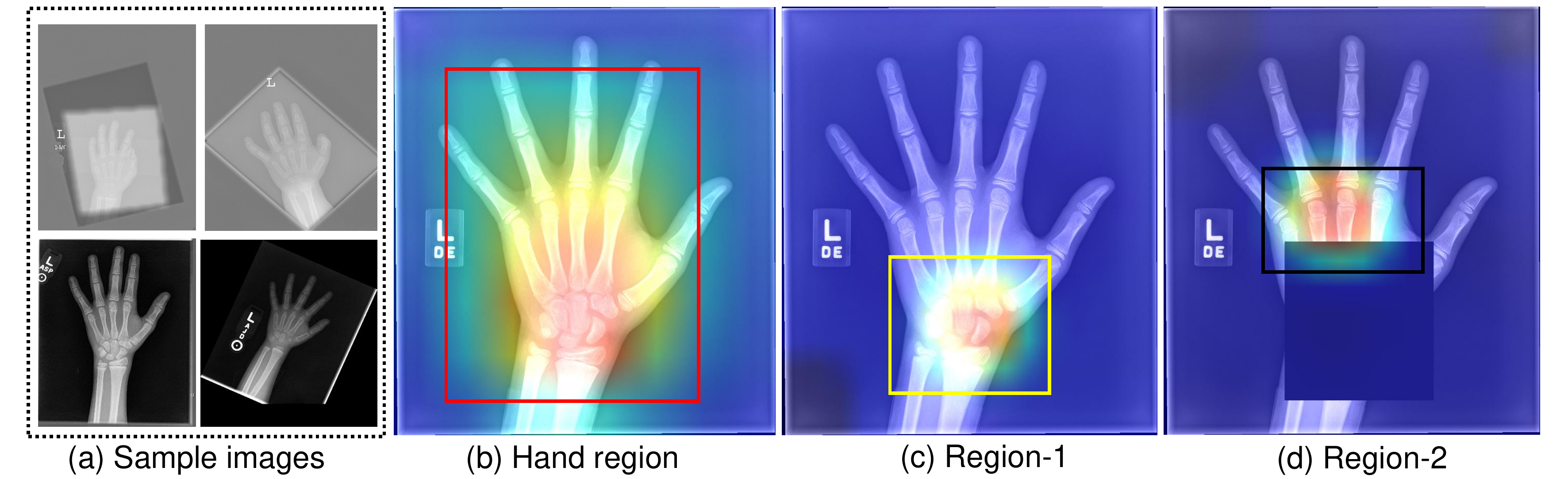}
\end{center}
\caption{Illustration of sample images from RSNA bone age dataset and the learned attention heat maps for the discriminative regions. (a) Four sample images selected from the RSNA pediatric bone age dataset. (b) The learned attention map for the hand region (H). (c) The learned attention map for the most discriminative region, which is denoted as Region-1 (R1). (d) The learned attention map for the next most discriminative region, which is denoted as Region-2 (R2).}
\label{fig1}
\end{figure}

In this work, we concentrate on deep learning  approaches for BAA. The difficulties of using deep learning for BAA are: (1) Raw input images are large (about $2000\times1500$ pixels), but bone age estimation is a fine-grained recognition task because ossification patterns are usually contained in specific small RoIs. Therefore, downsizing the raw images into low-resolution images will lose important information, decreasing the final performance. (2) Raw images can be poorly aligned. As shown in Fig. \ref{fig1}(a), the RoIs can be very small with undetermined position, which also reduces model performance. Some recent deep learning-based approaches have proposed to improve the BAA performance by localizing the RoIs \cite{escobar2019hand} or performing image alignment \cite{iglovikov2018paediatric} before age regression. Even though these methods demonstrated great performance improvement on BAA tasks, they suffer from two main limitations:
\begin{itemize}
\item In order to locate the informative local patches for BAA, most of these methods require the identification of RoIs or key points that are important for BAA and provide extra annotations for training \cite{iglovikov2018paediatric,escobar2019hand}. However, existing BAA datasets only contains image-level labels, and manually drawing RoIs and providing annotations can be subjective and expensive, and also require domain knowledge from expert radiologists.
\item Existing methods take BAA task as a general regression or classification problem, which uses mean absolute error ($\ell_1$ loss) or mean square error ($\ell_2$ loss), to penalize the differences between the estimated ages and the ground-truth ages. However, due to the label ambiguity problem in the age label space \cite{geng2016label}, this kind of loss function defined only based on a single label is suboptimal. Learning with a single label does not exploit the ordinal relationship among hand images with different individual ages, and leads to over-confident prediction.
\end{itemize}

To address these limitations, we present a novel attention guided deep learning framework for bone age expectation regression. Instead of downsizing the input images or training a detection or segmentation model using extra annotations, we propose to utilize attention maps to localize the most discriminative regions for BAA. Then, we aggregate different RoIs for both bone age expectation regression and age distribution learning. Our contribution are summarized as: (1)  As shown in Fig. \ref{fig1}, our method uses attention maps learned by deep models to automatically identify the hand region, the most discriminative region, and the next most discriminative region. It is also the first analysis to demonstrate systematically that the carpal and metacarpal bones are the two most important regions for BAA. (2) In order to leverage the correlation relationship between different individual ages and prevent the network from over-estimating classification confidence, we propose a joint age distribution learning and bone age expectation regression, which consistently improves performance. (3) By leveraging attention-guided local information and age distribution learning, our approach achieves competitive results without requiring manual annotations.

\begin{figure*}[!t]
\begin{center}
\includegraphics[width=0.75\linewidth]{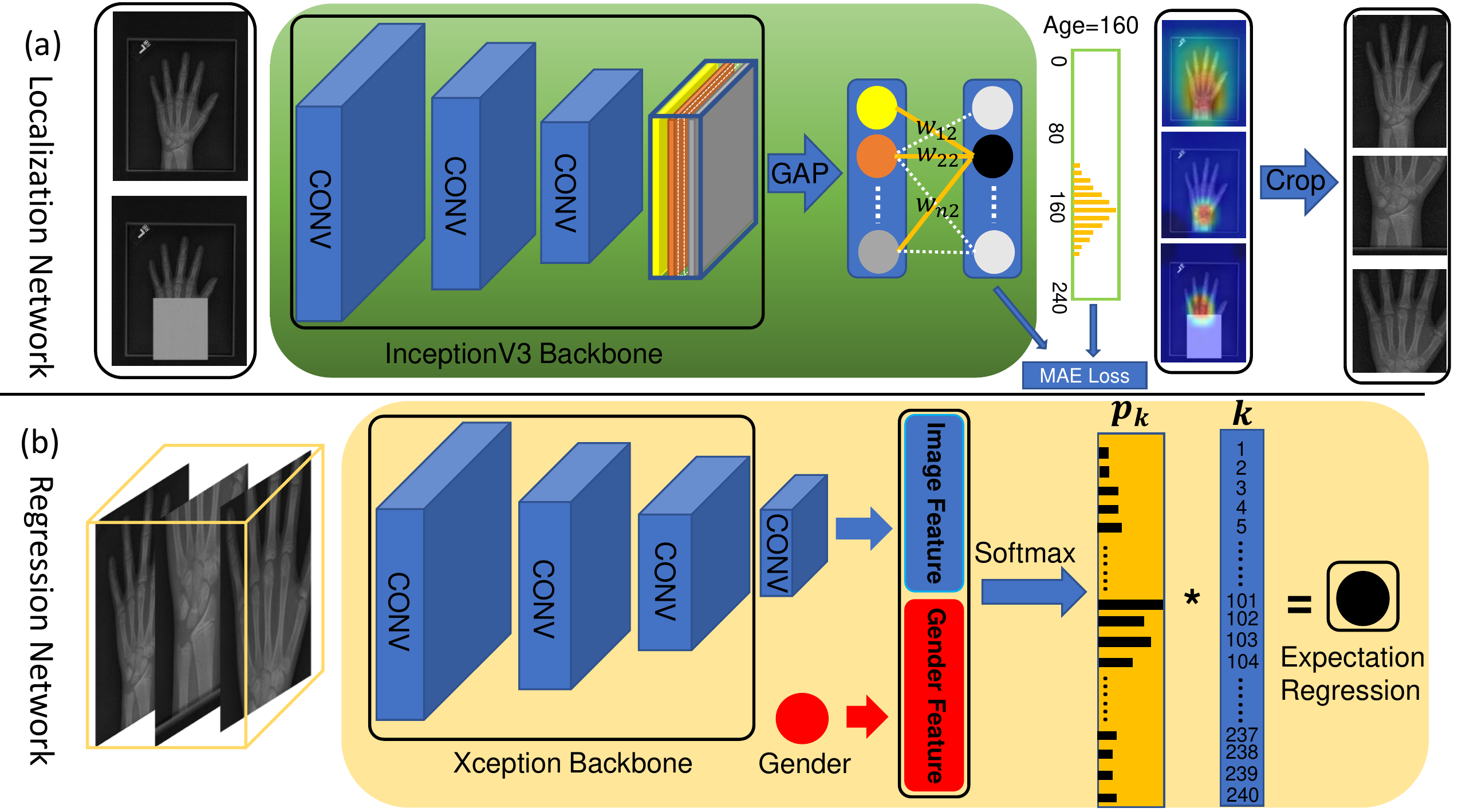}
\end{center}
\caption{Our approach consists of two stages. In the first stage, we train a classification model to learn the attention maps for the hand region, the most discriminative region, and the next most discriminative region. Guided by these attention maps, we crop the high resolution RoIs from the original image. In the second stage, we train a regression model to perform joint age distribution learning and expectation regression by aggregating different RoIs.}
\label{fig2}
\end{figure*}

\section{Related Work}
\textbf{Bone Age Assessment} Over the past decades, numerous automated image analysis methods and tools have been developed for BAA. These methods can be divided into two groups: non-deep learning based methods \cite{pietka2003integration,gertych2007bone,thodberg2008bonexpert,giordano2010automatic} and deep learning based methods \cite{halabi2019rsna,spampinato2017deep,lee2017fully,iglovikov2018paediatric,escobar2019hand}. Early representative non-deep learning-based methods mainly extract hand-designed features from the whole images or specific RoIs, and then train a classifier with no more than 2,000 samples. The performance of these methods is quite limited, with results ranging from 10-28 months mean absolute difference (MAD) \cite{spampinato2017deep}. Deep CNNs \cite{goodfellow2016deep,litjens2017survey} and a large scale BAA data set introduced by the Radiological Society of North America (RSNA) \cite{halabi2019rsna} have enabled recent advances to achieve impressive performance, with some exceeding an expert's performance \cite{cicero2017hand,halabi2019rsna,escobar2019hand}. Specifically, BoNet \cite{spampinato2017deep} designed an ad-hoc CNN for BAA, the author exploited the deformation layer to address bone nonrigid deformation, and achieved a result of 9.5 months MAD on average. In \cite{iglovikov2018paediatric}, in order to crop specific local regions, the author first trained an U-Net model to segment the hand region with 100 labeled hand masks and then trained a key point detection model to achieve image registration. As a result, they achieved a 6.30 months MAD for males and 6.49 months MAD for females. The winners of the RSNA challenge \cite{cicero2017hand} achieved a 5.99 months MAD with their best model and achieved a 4.26 months MAD by averaging 50 predictions (utilizing 5 top models with 10 augmented images). In the current best performing method, \cite{escobar2019hand} presented a new framework based on a local analysis of anatomical RoIs, the author provided extra bounding boxes and key point annotations during training, and performed hand detection and hand pose estimation to exploit local information for BAA. As a result, they achieved the best result using the RSNA bone age dataset, 4.14 months MAD.

\textbf{Attention Guided Part Localization}
Previous work mainly focuses on leveraging the extra annotations of bounding box and key point annotations to localize significant regions for bone age assessment\cite{iglovikov2018paediatric,escobar2019hand}. However, the heavy involvement of manual annotations and domain knowledge make it not practical in large-scale application scenarios. Recently, there have been numerous emerging studies working on attention guided localization, which allows the deep networks to focus on the informative task-relevant regions of the input images in an unsupervised manner\cite{zhou2016learning,selvaraju2017grad,zhang2018adversarial,fu2017look,zheng2017learning,guan2018diagnose}. Class activation mapping (CAM) \cite{zhou2016learning} revisits the global average pooling layer to enable the convolutional neural network (CNN) to be able to localize the discriminative image regions. In \cite{selvaraju2017grad}, Grad-CAM was proposed which is a generalization of CAM and is applicable to a significantly broader range of CNN model families. In \cite{fu2017look}, Fu et al. propose RA-CNN which recursively learns discriminative region attention and region-based feature representation at multiple scales for fine-grained image recognition. In the medical image analysis community, \cite{cai2018iterative} Cai et al. propose an attention mining (AM) strategy to improve the model’s sensitivity to disease patterns on chest X-ray images. In \cite{guan2018diagnose}, the authors propose an AG-CNN model, which enables the network to learn from disease-specific regions to avoid noise and improve alignment for thorax disease classification in chest X-ray images. Li et. al \cite{li2019attention} propose an attention-based multiple instance learning model for slide-level cancer grading and weakly-supervised RoI detection. Yang et. al \cite{yang2019guided} propose to use region-level supervision for the classification of breast cancer histopathology images, where the RoIs are localized and used to guide the attention of the classification network.

\section{Methodology}
As shown in Fig. \ref{fig2}, our proposal consists of two phases: an attention guided localization phase and a bone age expectation regression phase. In the localization phase, we train a classification model to learn the attention heat maps for the hand region, the most discriminative region, and the next most discriminative region. Guided by these attention maps, we then crop those high-resolution local patches from the original image. In the expectation regression phase, we train a regression model for joint age distribution learning and age expectation regression. The expectation regression model can exploit a single informative local patch or aggregate different local patches for BAA.

\subsection{Phase I: Attention Guided RoIs Localization}
Weakly supervised detection and localization methods that aim to identify the location of the object in a scene only using image-level labels have been widely used for many vision tasks \cite{zhou2016learning,selvaraju2017grad,zhang2018adversarial} and medical image analysis \cite{guan2018diagnose,yang2019guided}. Inspired by these methods, we propose to utilize learned attention maps to identify the discriminative local patches for BAA. As shown in Fig. \ref{fig2}(a), for a given CNN model and an input image, let $\bm{F}\in R^{H\times W\times C}$ denote the activation outputs of the last convolutional layer. The resulting feature maps are then fed into a global average pooling (GAP) or global max pooling (GMP) layer \cite{zhou2016learning}, followed by a fully connected (FC) layer. For convenience, we only consider the case of using the GAP layer and ignore the bias term. We denote the average value of the $k$-th feature map as $S_k=\tfrac{\sum_{i,j}F_{ijk}}{H\times D}, k=0,1,\cdots,C-1$, and denote the weight matrix of the FC layer as $\bm{W}\in R^{C\times T}$, where $T$ is the number of classes in the classification model. In this way, the value of the $t$-th output node can be calculated as
\begin{equation}\label{eq1}
\hat{\bm{Y}_t}=\sum_{k=0}^{C-1}\bm{W}_{kt}S_k=\frac{1}{H\times W}\sum_{i,j}\sum_{k=0}^{C-1}\bm{W}_{kt}\bm{F}_{ijk}
\end{equation}
where $\hat{\bm{Y}}\in R^T$ is the network output and $\bm{W}_{kt}$ denotes the connection weights between the $k$-th input nodes and $t$-th output nodes in the last FC layer. Therefore, for the $t$-th class samples, we define a heat map $\bm{A}_t\in R^{H\times W}$ as,
\begin{equation}\label{eq2}
\bm{A}_t(i,j)=\sum_{k=0}^{C-1}\bm{W}_{kt}\bm{F}_{ijk}
\end{equation}
The final output of $t$-th node, therefore, can be calculated as
\begin{equation}\label{eq3}
\hat{\bm{Y}_t}=\frac{1}{H\times W}\sum_{i,j}\bm{A}_t(i,j)
\end{equation}
In this respect, for a given image that is assigned to class $t$, the heat map $\bm{A}_t$ indicates the contribution of each pixel to the final classification result. After obtaining the heat map $\bm{A}_t$, we resize the it to the original image size and design a binary mask $\bm{M}_t$ to identify the most discriminative regions of a given image.
\begin{equation}\label{eq4}
\bm{M}_t(i,j) =
\begin{cases}
1 & \bm{A}_t(i,j)\geq\tau \\
0 & \bm{A}_t(i,j)<\tau
\end{cases}
\end{equation}
 where $\tau$ is a threshold that determines the size of the RoIs. A larger $\tau$ leads to a smaller RoI, and vice versa. Guided by the binary mask, we can crop the high-resolution discriminative local patches from the original images.

\textbf{Implementation Details} For the classification model, we adopt the InceptionV3 (without top layers) as the backbone network for feature extraction, and then add a GMP (or GAP) layer followed by a FC layer with 240 output nodes, which is the maximum age of the children in the data set in months. When we utilize the original one-hot labels for training, the network fails to converge. We believe the reason is that hand images with different ages are similar, but have different one-hot labels. Hence, we utilize soft labels for training. For a hand image and its labeled age $t$, we define the following function to soften the label distribution
\begin{equation}\label{eq5}
\bm{Y}_i =
\begin{cases}
1-\frac{\vert i-t \vert}{l} & \vert i-t \vert \leq l \\
0 & \vert i-t \vert > l
\end{cases}
\end{equation}
where $\bm{Y}\in R^T$ is the ground-truth label distribution and $i=1,2,\cdots,240$. $l$ controls the smoothness of the label distribution, a larger $l$ leads to a smoother label distribution. In the experiments, we set $l=50$. We utilize the weights pre-trained in ImageNet, and train the network with the Adam optimizer with a batch size of 32. The network is trained over 70 epochs, the learning rate is set to 0.0003 for the first 50 epochs and set to 0.0001 for the last 20 epochs.

\noindent\textbf{Localization of Region-1}
To localize the most discriminative region (Region-1), we train the classification model with the original images which have been resized to $560\times560$. The activation outputs $\bm{F}\in R^{18\times18\times2048}$ are then fed into a GMP layer which follow by the last FC layer. The localization of the Region-1 can therefore, be given by the binary mask in Eq. \ref{eq4}, where the threshold is set to $\tau=50$ empirically.

\begin{figure*}[!t]
\begin{center}
\includegraphics[width=0.92\linewidth]{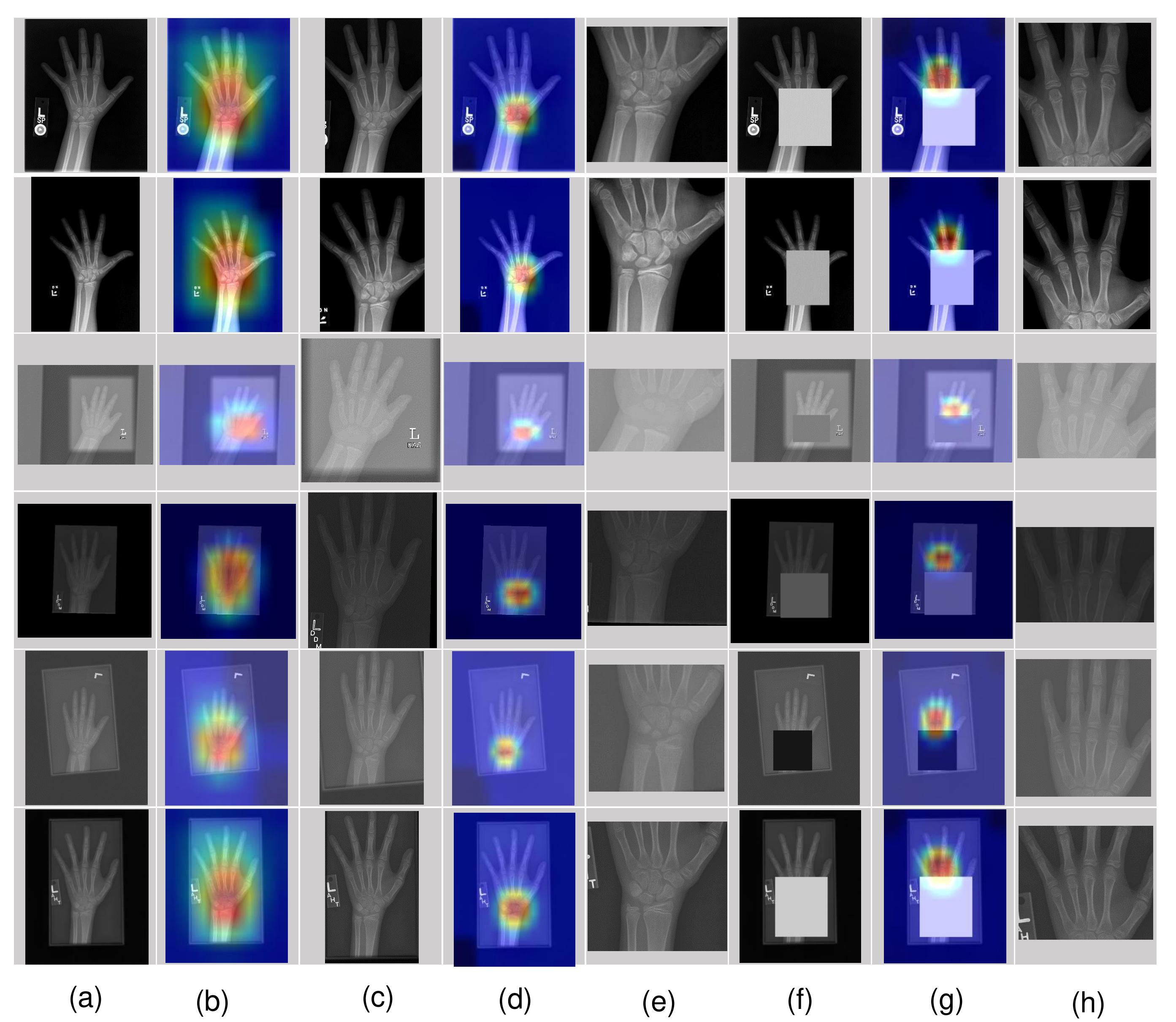}
\end{center}
\caption{Six representative samples with different ages. (a) Original images (O), (b) images with attention map for hand region, (c) the cropped hand region (H), (d) images with attention map for Region-1, (e) the cropped Region-1 (R1), (f) the images with pixels in Region-1 erased (E), (g) images with attention map for Region-2, (h) the cropped Region-2 (R2)}
\label{fig3}
\end{figure*}

\noindent\textbf{Localization of Region-2}
To localize the next most discriminative region (Region-2), we generate input images by replacing the pixels in Region-1 with random values. As shown in Fig. \ref{fig3}(f), training the classification network using the images with Region-1 "erased" forces the network to make predictions based on the pixels other than those in Region-1. In this way, we can localize Region-2 in the same way as localizing Region-1.

\noindent\textbf{Localization of Hand Region} The introduced method tends to localize a small discriminative task-relevant region \cite{zhang2018adversarial}. To make the attention heat maps focus on the whole hand region, we utilize a smaller input image by resizing the original images to $300\times300$. In this way, each pixel in the feature map $\bm{F}\in R^{8\times8\times2048}$ will correspond to a larger image patch in the original image. We also use the GAP layer instead of the GMP layer, which also helps to concentrate on a larger discriminative region. We empirically set the threshold $\tau=20$ to obtain the attention maps for the full hand.

\subsection{Phase II: Bone Age Expectation Regression}
\textbf{Network Design} In the second phase, we perform bone age expectation regression with the high-resolution local patches. The different local patches are aggregated by feeding into different input channels. As shown in Fig. \ref{fig2}(b), we adopt the Xception \cite{chollet2017xception} without top layers as the backbone network, followed by a convolutional layer, a max pooling layer, and a FC layer. To effectively utilize gender information, we concatenate the image features with the gender features, which takes gender information (1 for male and -1 for female) as input and feeds it through a FC layer with 32 neurons. The concatenated features are then fed into the last FC layer with softmax activation. The softmax output $p_k, k = \{1,2,\cdots,240\}$ represents the bone age distribution (the probability of belonging to different ages), which is used to calculate the expectation of bone age.

\textbf{Joint Age Distribution Learning for BAA} Hand X-ray images look very similar if the age of these images are close. For example, one's hand X-ray image looks the same when he is 160 or 161 month. This inspire us to make use of the correlation information of the hand images at neighboring ages. However, existing approaches take the BAA task as a regression problem or discrete classification problem, which can not exploit the correlation information among neighboring ages. Inspired by the label distribution learning (LDL) \cite{geng2016label,geng2013facial,huo2016deep,gao2018age}, we propose to learn an age distribution rather than a single age label for each hand image. The age distribution contains a group of probability values which represent the degree of each age to the hand image. It also reflects the ordinal relationship among neighboring ages. Formally, let $\bm{x}_i$ and $g_i\in\{-1,1\}$ denotes the local patches and gender indicator of $i$-th sample, $y_i\in\{1,2,\cdots,240\}$ denotes the corresponding label.  As shown in Fig. \ref{fig2}(b), we assume that $F(\bm{x}_i)\in R^m$ is the image feature and $G(g_i)\in R^n$ is the gender feature for $i$-th sample, where $m$ and $n$ are dimension of the image and gender feature. We fuse the image and gender information by concatenation $\bm{f}_i=[f(\bm{x}_i);G(g_i)]\in R^{m+n}$, followed by a full connected layer which transfers $\bm{f}_i$ to $\bm{z}_i\in R^{240}$ by
\begin{equation}\label{eq6}
\bm{z}_i = \bm{W}^{\top}\bm{f}_i + \bm{b}
\end{equation}
Then, we employ a softmax activation function to turn $\bm{z}_i$ into the age distribution,
\begin{equation}\label{eq7}
p_i^k = \frac{\exp(\bm{z}_i^k)}{\sum_k\exp(\bm{z}_i^k)}, \quad k=1,2,\cdots,240
\end{equation}
here $\bm{p}_i\in R^{240}$ denotes the age distribution of $i$-th sample and $p_i^k, k=\{1,2,\cdots,240\}$ denotes the probability that the $i$-th hand image belongs to age $k$ (month). Finally, the output layer take the age distribution and label set as input and output the expectation of bone age
\begin{equation}\label{eq8}
\hat{y}_i = \sum_{k=1}^{k=240}k\cdot p_i^k
\end{equation}
Given input local patches and corresponding gender information, the regression model aims to minimize the mean absolute error between the ground truth ages $y$ and the estimated expectation ages $\hat{y}$
\begin{equation}\label{eq9}
\ell_{MAE} = \sum_i\Vert y_i-\hat{y}_i\Vert_1
\end{equation}
Intuitively, the estimated age distribution $\bm{p}$ should be concentrated at a small range of the ground truth age, and always follow a Gaussian distribution \cite{geng2016label}. However, this property can not be guaranteed with the MAE loss between ground truth and age expectation. To obtain an reasonable age distribution and prevent the expectation regression model from over-estimation of confidence \cite{muller2019does}, we add a regularization term to minimize the Kullback-Leibler (KL) divergence between the estimated age distribution $\bm{p}_i$ and a Gaussian distribution $\bm{G}_i$.
\begin{equation}\label{eq10}
\ell_{reg} = \sum_iD_{KL}(\bm{p}_i\|\bm{G}_i) = -\sum_i\sum_{k=1}^{240}G_i^k\ln\frac{p_i^k}{G_i^k}
\end{equation}
where $\bm{G}_i$ is a Gaussian distribution generated from ground truth age $y$ and a hyper-parameter $\delta$ which controls the sharpness of the Gaussian distribution.
\begin{equation}\label{eq10}
G_i^k = \frac{1}{\sqrt{2\pi}\delta}\exp(-\frac{(k-y_i)^2}{2\delta^2}), \quad k=1,2\cdots,240
\end{equation}
The final loss function minimizes the MAE loss and regularization loss jointly by
\begin{equation}\label{eq11}
\ell = \ell_{MAE} + \lambda\ell_{reg}
\end{equation}
where $\lambda$ balances the contribution of the expectation regression loss and regularization loss. Compared with general regression model with $\ell_1$ loss, our proposed method regresses the bone age expectation and learns the age distribution implicitly. Learning an age distribution rather than a single age label will enable our model make use of the ordinal relationship among different individual ages, and prevent the network from over-confidence which leads to more robust age estimation.

\begin{table*}[ht]
\centering
\caption{mIoU and mAP performance of the attention guided RoIs localization}
\label{tab1}
\begin{tabular}{|c|c|c|c|c|c|c|c|}
\hline
Threshold & 10   & 20    &30    & 40    & 50    & 60    & 80    \\ \hline
\hline
Hand      & 0.576/0.780 & 0.757/0.995  &0.710/0.980 & 0.695/0.970 & 0.666/0.920 & 0.633/0.840 & 0.565/0.685 \\ \hline
Carpal Bones  & 0.170/0.0 & 0.223/0.125  &0.545/0.635  & 0.679/0.880 & 0.722/0.965 & 0.700/0.945 & 0.645/0.870 \\ \hline
Metacarpal Bones  & 0.112/0.0 & 0.274/0.245  &0.521/0.605  & 0.541/0.695 & 0.560/0.735 & 0.565/0.735 & 0.553/0.730 \\ \hline
\end{tabular}
\end{table*}

\begin{table*}[ht]
\centering
\caption{Test results (MAE/month) with different baseline networks without and with pre-training}
\label{tab2}
\begin{tabular}{|c|>{\centering\arraybackslash}p{1cm}|>{\centering\arraybackslash}p{1cm}|>{\centering\arraybackslash}p{1cm}|>{\centering\arraybackslash}p{1cm}|>{\centering\arraybackslash}p{1cm}|>{\centering\arraybackslash}p{1cm}|>{\centering\arraybackslash}p{1cm}|>{\centering\arraybackslash}p{1cm}|}
\hline
Network           & \multicolumn{2}{c|}{Vgg19} & \multicolumn{2}{c|}{InceptionV3} & \multicolumn{2}{c|}{ResNet50} & \multicolumn{2}{c|}{Xception} \\ \hline\hline
w/o \& w pre-training & 12.2         & 9.3         & 10.9            & 9.2            & 11.3           & 9.3          & 9.9           & 8.8           \\ \hline
\end{tabular}
\end{table*}

\textbf{Implementation details} Our model was implemented with keras 2.1.6 and trained on NVIDIA DGX system with 8 NVIDIA V100 GPUs and 512G memory. The weights of the backbone network are initialized with weights pre-trained on ImageNet, and we train the network with the Adam optimizer using a batch size of 16. The network is then trained for 120 epochs. The learning rate is set to 0.0003, 0.0001, and 0.00001 for the first 60 epochs, the next 30 epochs, and the final 30 epochs, respectively. Optimal hyper-parameters are determined using a grid search strategy. The best threshold for the attention map is selected from $\tau\in\{10,20,\cdots,100\}$ and the best trade-off parameter is selected from $\lambda\in\{0,0.001,0.01,0.05,0.1,0.5,1.0,5.0\}$, the parameter sensitivity analysis can be seen in Fig. \ref{fig5}. Since the performance is not sensitive to the variance of the given Gaussian distribution, we empirically set it to $\delta=15$ throughout the experiments.

\section{Experiments}
\subsection{Dataset and Evaluation Protocol}
\textbf{Dataset} We evaluate the performance of our approach using the dataset from the 2017 Pediatric Bone Age Challenge organized by the Radiological Society of North America (RSNA) \cite{halabi2019rsna}. The dataset is freely available now and can be download at \footnote{https://www.kaggle.com/kmader/rsna-bone-age}. The example hand images can be seen in Fig. \ref{fig3}(a). All the hand images are of arbitrary size (about $2000\times1500$) and each image contains bone age (1-240) and gender information (0 and 1 for male and female). During the experiments, we randomly split the dataset into three splits, with 500 samples each for validation and testing, and the remaining images used for training.

\textbf{Evaluation metrics} We take mean absolute error (MAE) as loss function and main evaluation criteria throughout the experiments. The quantitative performance of attention guided localization was measured by mean Intersection over Union (mIoU) and Average Precision ($\text{AP}_{50}$). $\text{IoU}=\frac{\text{area of overlap}}{\text{area of union}}$ reflects the overlap ratio between the ground truth bounding box and predicted bounding box. mIoU measures the mean IoU in the test set. $\text{AP}_{50} = \frac{\#(\text{IoU}>0.5)}{N}$ represents the average precision of region localization, where $\#(\text{IoU}>0.5)$ denotes the number of test samples with $\text{IoU}>0.5$, and $N$ is the number of test samples.

\subsection{Evaluation of the RoI Localization}
\noindent\textbf{Visualization} As shown in Fig. \ref{fig3}, to demonstrate the effectiveness of the attention-guided discriminative region localization, we show six representative images and their corresponding attention maps and cropped local patches. Fig. \ref{fig3} reveals several interesting observations: (1) Although the hand region in the original images are in various angles and arbitrarily-sized, the learned attention maps can always localize the hand region accurately. (2) The carpal bones are identified as the most informative and discriminative local regions, which is consistent with the manually determined RoIs in the TW-based method \cite{spampinato2017deep}. (3) The joints of the metacarpal bones are recognized as the next most discriminative regions, which is also consistent with RoIs marked by radiologists \cite{iglovikov2018paediatric}. (4) Compared to the original images, the hand region, Region-1, and Region-2 are better aligned across different hand images.

\noindent\textbf{Quantitative Evaluation} To demonstrate that the introduced attention guided method can always locate the full hand region, carpal bones, and metacarpal bones, we created manual bounding box labels for the full hand, carpal bones, and metacarpal bones for 200 images randomly selected from the test set. RoI regions for the hand, carpal bones, and metacarpal bones were then automatically determined by thresholding the learned attention maps, and the performance of region localization was measured by mIoU and $\text{AP}_{50}$ score. As observed in Table \ref{tab1}, we adjusted the threshold $\tau$ which resulted in different localization performances: (1) For localization of the full hand region, the mIoU was 0.757 and the $\text{AP}_{50}$ was $0.995$ when the threshold was set to $\tau=20$. (2) For the localization of Region-1 (carpal bones), the maximum mIoU and $\text{AP}_{50}$ are 0.722 and 0.965 respectively when the threshold is set as $\tau=50$. (3) For Region-2 (metacarpal bones), the maximum mIoU and $\text{AP}_{50}$ are $0.565$ and $0.735$ respectively. (4) Note that the best threshold for Hand region is quite different from the threshold for Region-1 and Region-2, this is because we use a smaller input image to get a larger attention map for the full hand region. The results show that all three regions consistently overlap with the manually segmented areas. Specifically, the hand region and the carpal bones can be located accurately when selecting an appropriate threshold. For the metacarpal bones, the mIoU and $\text{AP}_{50}$ scores are much worse which shows the localization is less accurate than the hand region and carpal bones. We believe the reason is that the Region-2 is less discriminative and the ground truth for Region-2 is more subjective. It is worth noting that the ground truth bounding box of the Region-1 and Region-2 are subjective and the regions are small in size, so it is not surprising that the mIoU values are not close to perfect. Besides, the goal of the RoI generation step is to locate and aggregate the most discriminatvie local region for bone age assessment (BAA), which is a different task from region localization. The mIoU and $\text{AP}_{50}$ scores only provide a rough assessment of RoI generation, but the true assessment of the quality of the RoI localization is reflected by the accuracy of BAA using a single local region, which is given in Table \ref{tab5}.

\begin{table*}[ht]
\centering
\caption{Experimental results with and without using gender information (Xception as backbone), different fully connected (FC) layer architectures are used to incorporate the gender information. 1$\rightarrow$16 represents taking gender information (1 for male and -1 for female) as input and feeds it through a FC layer with 16 neurons}
\label{tab3}
\begin{tabular}{|>{\centering\arraybackslash}p{1.8cm}|>{\centering\arraybackslash}p{1.8cm}|>{\centering\arraybackslash}p{1.8cm}|>{\centering\arraybackslash}p{1.8cm}|>{\centering\arraybackslash}p{1.8cm}|>{\centering\arraybackslash}p{1.8cm}|}
\hline
Architectures  & w/o & 1$\rightarrow$16 & 1$\rightarrow$32 & 1$\rightarrow$64 & 1$\rightarrow$64$\rightarrow$32  \\ \hline\hline
MAE         & 8.8     & 7.9     & 7.8     & 7.8   &8.0  \\ \hline
\end{tabular}
\end{table*}

\begin{table*}[ht]
\centering
\caption{Experimental results with different image sizes using gender information (Xception as backbone)}
\label{tab4}
\begin{tabular}{|>{\centering\arraybackslash}p{1.8cm}|>{\centering\arraybackslash}p{1.8cm}|>{\centering\arraybackslash}p{1.8cm}|>{\centering\arraybackslash}p{1.8cm}|>{\centering\arraybackslash}p{1.8cm}|>{\centering\arraybackslash}p{1.8cm}|}
\hline
image size  & 224$\times$224 & 336$\times$336 & 448$\times$448 & 560$\times$560 & 720$\times$720  \\ \hline\hline
MAE         & 7.8     & 7.6     & 7.4     & 7.3   &7.3  \\ \hline
\end{tabular}
\end{table*}

\begin{table*}[]
\centering
\caption{Performance comparison between our proposal and state-of-the-art methods.}
\label{tab5}
\begin{tabular}{|c|l|c|c|c|c|c|c|c|c|c|l|}
\hline
\multicolumn{2}{|c|}{Methods} & \multicolumn{2}{c|}{Image Size} & \multicolumn{2}{c|}{Extra Labels} & \multicolumn{2}{c|}{Data Augment} & \multicolumn{2}{c|}{Model Ensembling} & \multicolumn{2}{c|}{MAE}     \\ \hline\hline
\multicolumn{2}{|c|}{\cite{iglovikov2018paediatric}}       & \multicolumn{2}{c|}{$750\times750$}           & \multicolumn{2}{c|}{mask \& keypoint}                  & \multicolumn{2}{c|}{Yes}                & \multicolumn{2}{c|}{18 model results}           & \multicolumn{2}{c|}{6.4}        \\ \hline
\multicolumn{2}{|c|}{\cite{spampinato2017deep}}       & \multicolumn{2}{c|}{224$\times$224}           & \multicolumn{2}{c|}{No}                  & \multicolumn{2}{c|}{Yes}                & \multicolumn{2}{c|}{No}           & \multicolumn{2}{c|}{9.5}        \\ \hline
\multicolumn{2}{|c|}{\multirow{2}{*}{\cite{cicero2017hand}}}       & \multicolumn{2}{c|}{$512\times512$}           & \multicolumn{2}{c|}{No}                  & \multicolumn{2}{c|}{Yes}                & \multicolumn{2}{c|}{No}           & \multicolumn{2}{c|}{5.99}        \\ \cline{3-12}
\multicolumn{2}{|l|}{}       & \multicolumn{2}{c|}{$512\times512$}           & \multicolumn{2}{c|}{No}                  & \multicolumn{2}{c|}{Yes}                & \multicolumn{2}{c|}{50 model results}           & \multicolumn{2}{c|}{4.26}        \\ \hline
\multicolumn{2}{|c|}{\cite{escobar2019hand}}       & \multicolumn{2}{c|}{500$\times500$}           & \multicolumn{2}{c|}{Bbox \& keypoint}                  & \multicolumn{2}{c|}{Yes}                & \multicolumn{2}{c|}{No}           & \multicolumn{2}{c|}{4.14}        \\ \hline\hline
\multicolumn{2}{|c|}{\textbf{Ours}}   & $\;\;$ O$\;\;$              & H              & $\quad$R1$\quad$                 & R2                & H+R1             & R1+R2             & O+H+R1         & H+R1+E         & \multicolumn{2}{c|}{H+R1+R2} \\ \hline
\multicolumn{2}{|c|}{$\ell_1$}    & 7.3            & 6.4            & 6.1                & 7.0               & 5.4              & 5.6               & 5.4            & 4.7            & \multicolumn{2}{c|}{4.8}     \\ \hline

\multicolumn{2}{|c|}{$\ell_{MAE}+\ell_{reg}$}    & 6.2            & 5.6            & 5.3                & 6.2               & 4.8              & 5.1               & 4.7            & 4.3            & \multicolumn{2}{c|}{4.3}     \\ \hline
\end{tabular}
\end{table*}

\subsection{Evaluation of Bone Age Assessment}
\noindent\textbf{Baseline networks}
In order to select a better baseline network for BAA task, we first train four widely used CNN networks, including VGG19, InceptionV3, ResNet50 and Xception network, to  perform bone age regression. We didn't use the gender information and set the input image size to $224\times224$. In order to study the effect of pre-training on final performance, we train the baseline networks (a) from scratch and (b) from the model pre-trained in ImageNet. The results are shown in Table \ref{tab2}. We observed that: (1) Whether using a pre-trained model or not, Xception network\cite{chollet2017xception} achieves the best performance among all the baseline networks, we believe this is because the Xception network uses the depth-wise separable convolution which reduces the number of free parameters and therefore prevents the network from over-fitting. (2) Initializing the model parameters with the weights pre-trained from ImageNet consistently improves the performance over all baseline networks.

\noindent\textbf{Gender information and input image size} Due to the physical differences between men and women, gender information is also an important factor for BAA. To investigate how gender information influences the final performance, we tried several different fully connected architectures to leverage the gender information. As shown in Table \ref{tab3}, utilizing the gender information improves the performance from $8.8$ to $7.8$ with Xception backbone network, and the performance is not sensitive to the gender feature embedding networks. Therefore, we utilize one FC layer with 32 neurons to leverage the gender information for BAA. Since bone age estimation is a very fine-grained recognition task, increasing the size of the input image can always improve performance, in order to determine the appropriate input image size, we train the Xception network with gender input under different input sizes. As observed in Table \ref{tab4}: Increasing the input image size from $224\times224$ to $560\times560$ improves performance from $7.8$ to $7.3$, but it does not help when the image size is larger than $560\times560$. Therefore, we set the input image size to $560\times560$ in the following experiments.


\begin{figure*}[!ht]
    \centering
  \begin{subfigure}[b]{0.24\textwidth}
    \includegraphics[width=\textwidth]{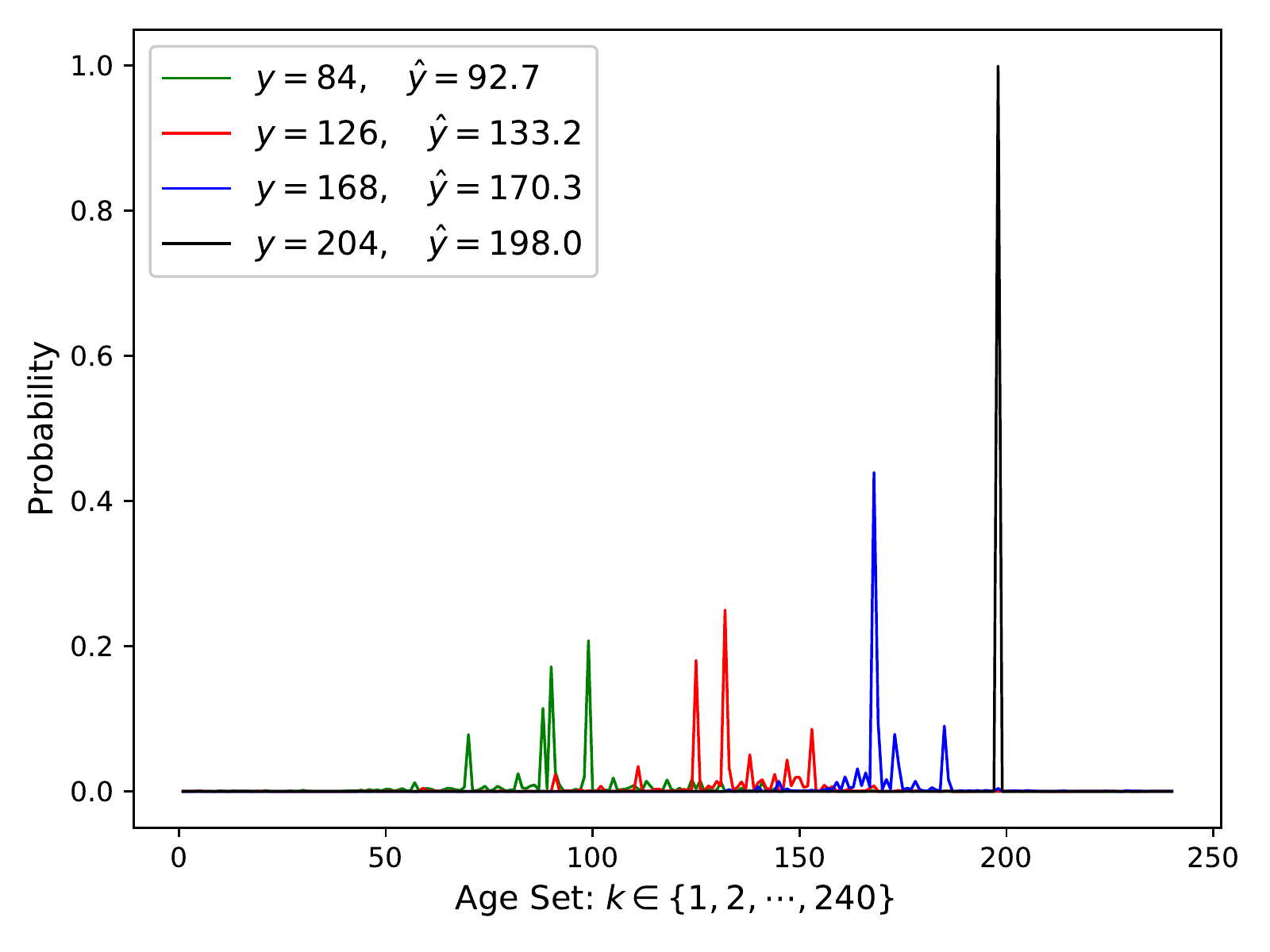}
    \caption{$\lambda=0$}
    \label{fig41}
  \end{subfigure}
    \begin{subfigure}[b]{0.24\textwidth}
    \includegraphics[width=\textwidth]{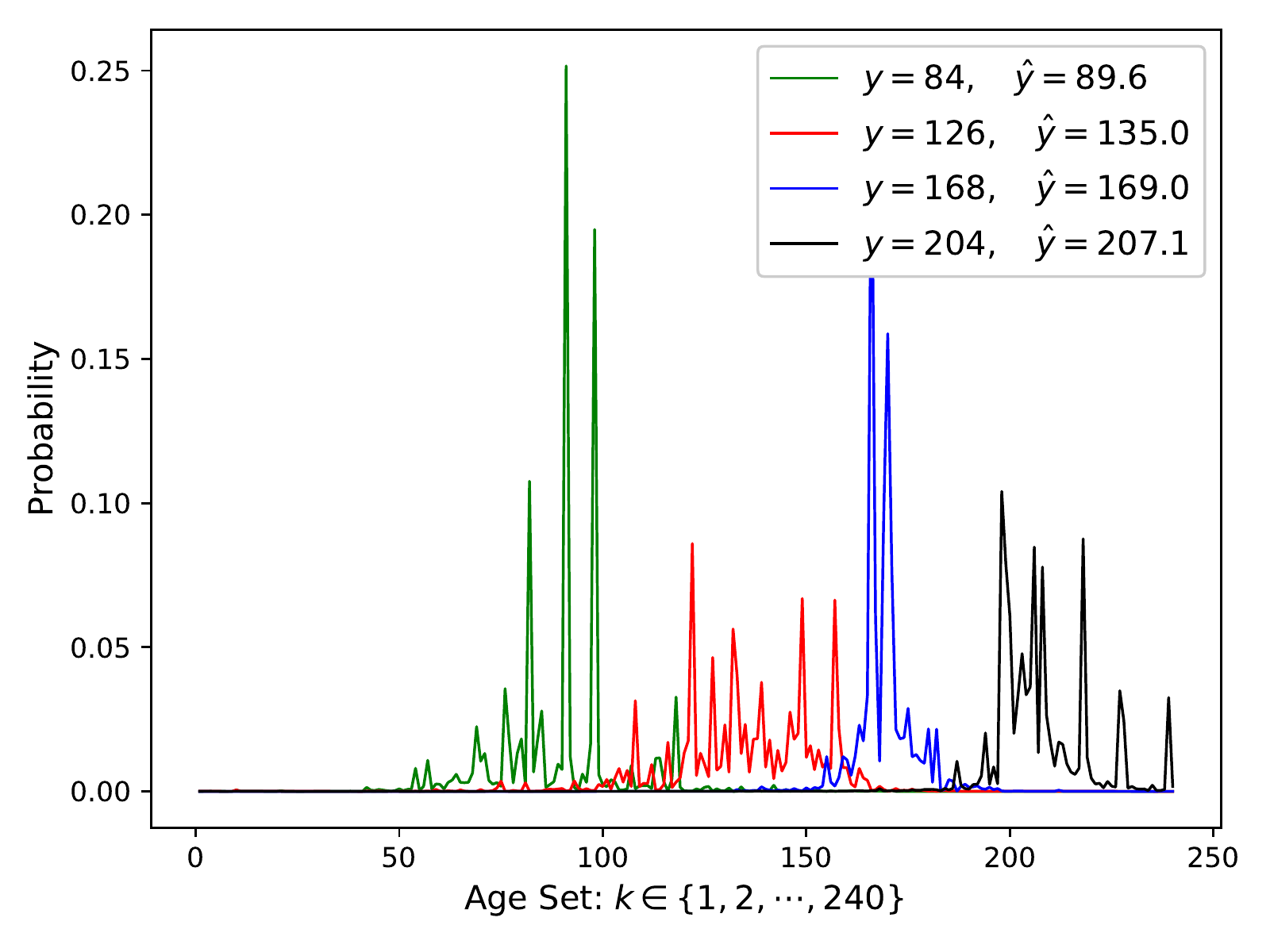}
    \caption{$\lambda=0.05$}
    \label{fig412}
  \end{subfigure}
   \begin{subfigure}[b]{0.24\textwidth}
    \includegraphics[width=\textwidth]{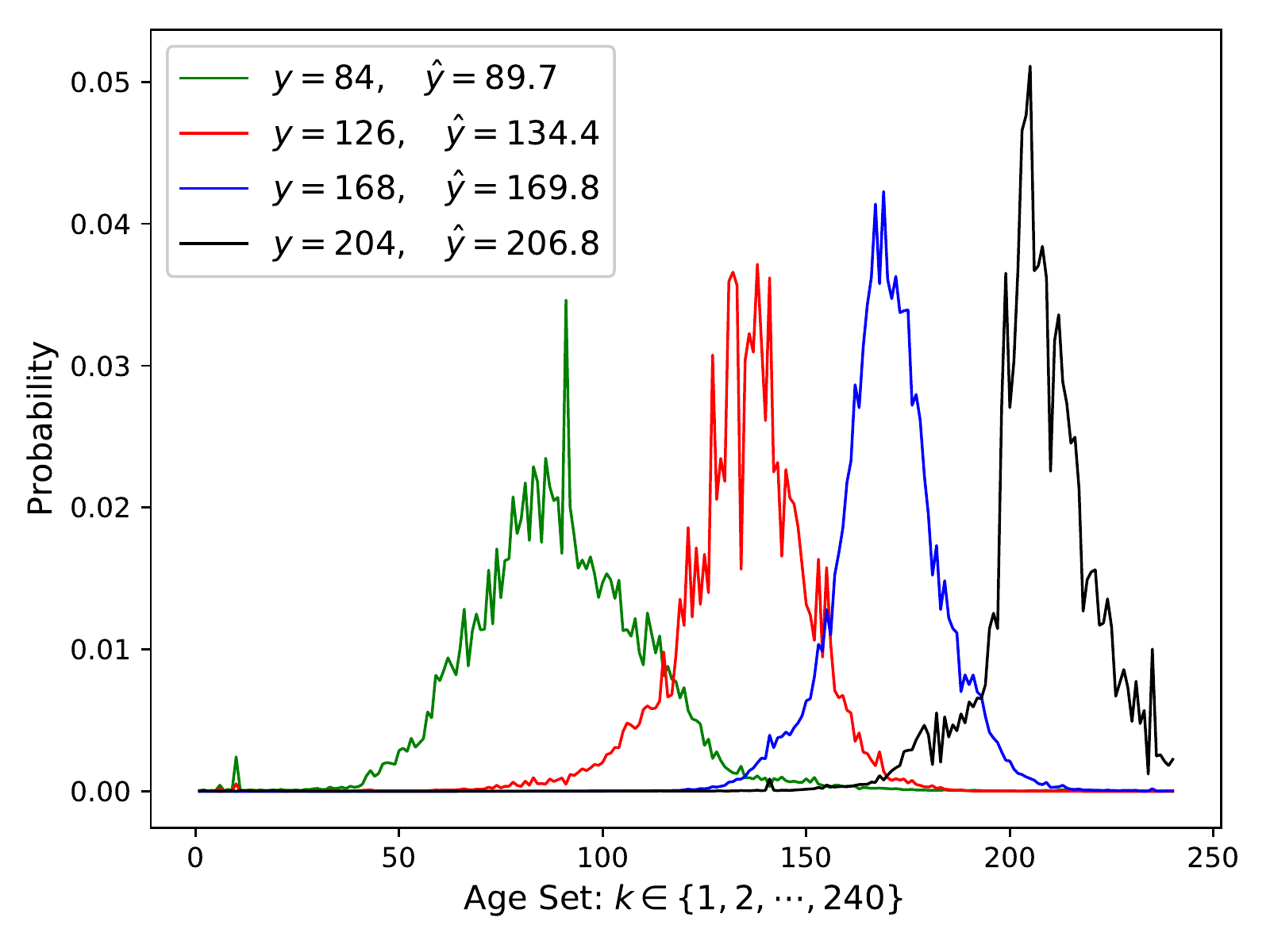}
    \caption{$\lambda=0.5$}
    \label{fig42}
  \end{subfigure}
  \begin{subfigure}[b]{0.24\textwidth}
    \includegraphics[width=\textwidth]{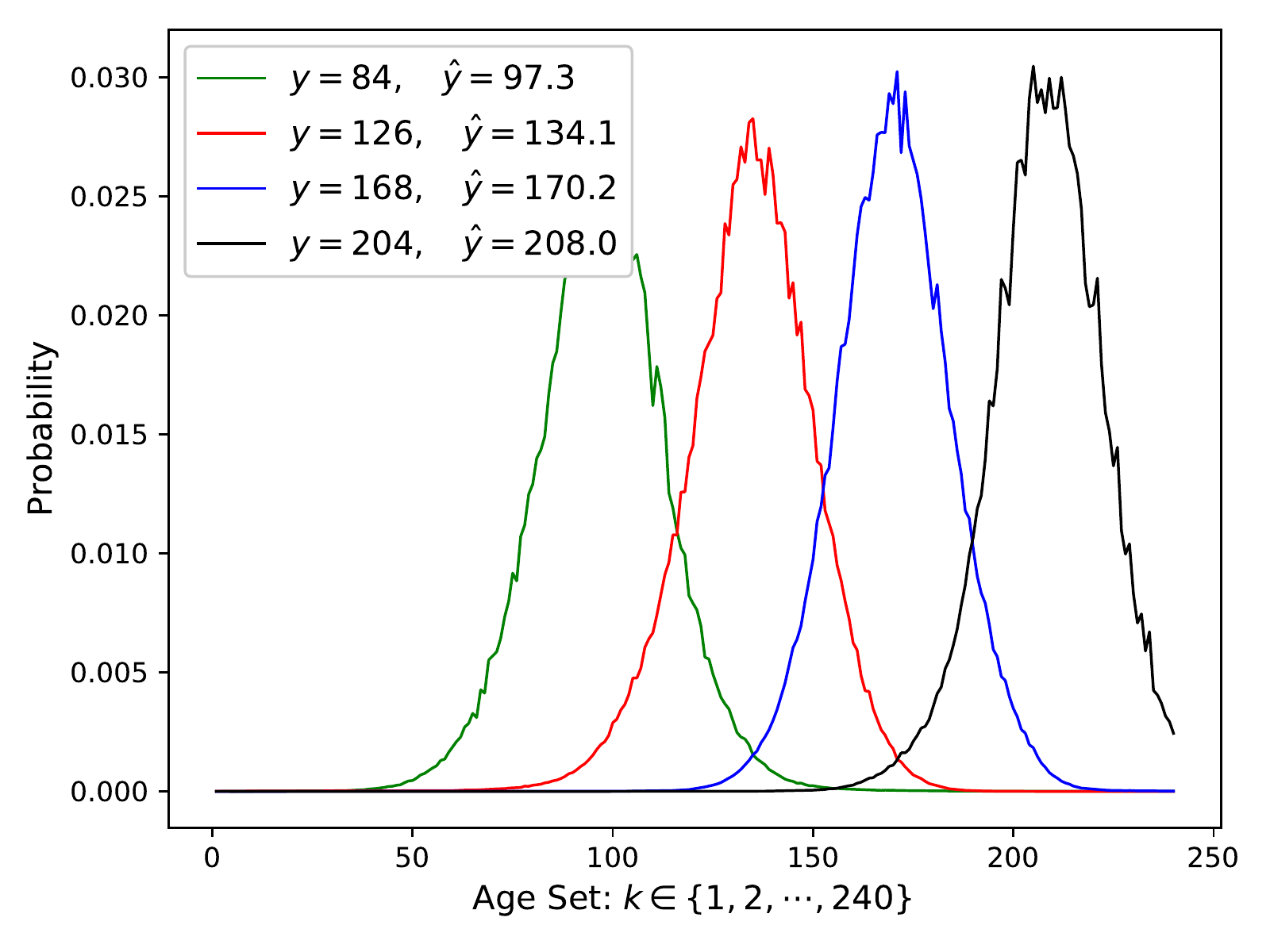}
    \caption{$\lambda=5$}
    \label{fig43}
  \end{subfigure}
\caption{Learned Age distribution for four test samples that distributed in different age range. The larger the trade-off parameter $\lambda$, the closer the age distribution is to Gaussian distribution.}
\label{fig4}
\end{figure*}

\begin{figure}
    \centering
  \begin{subfigure}[b]{0.24\textwidth}
    \includegraphics[width=\textwidth]{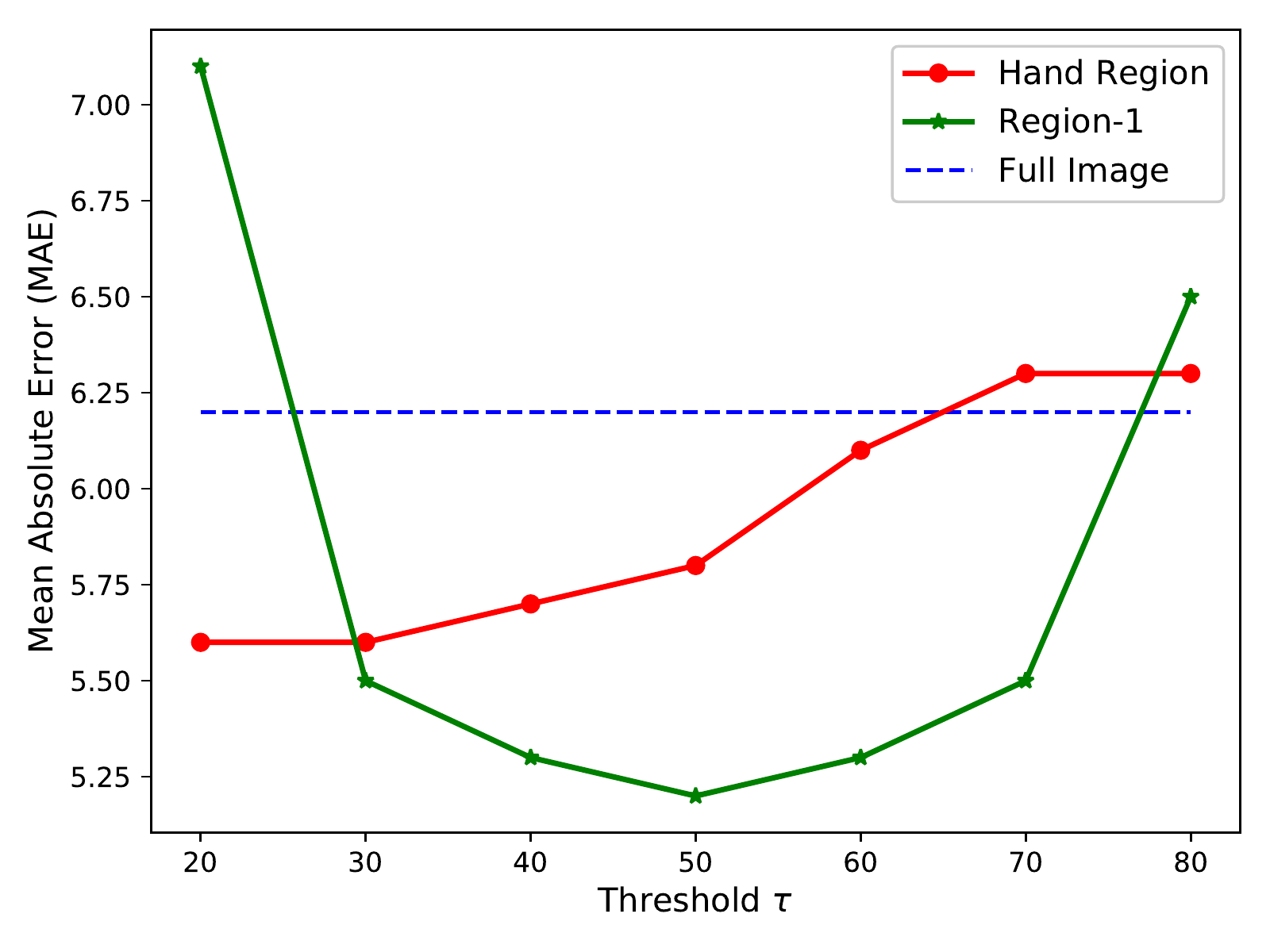}
    \caption{Parameter sensitivity w.r.t. $\tau$}
    \label{fig51}
  \end{subfigure}
    \begin{subfigure}[b]{0.24\textwidth}
    \includegraphics[width=\textwidth]{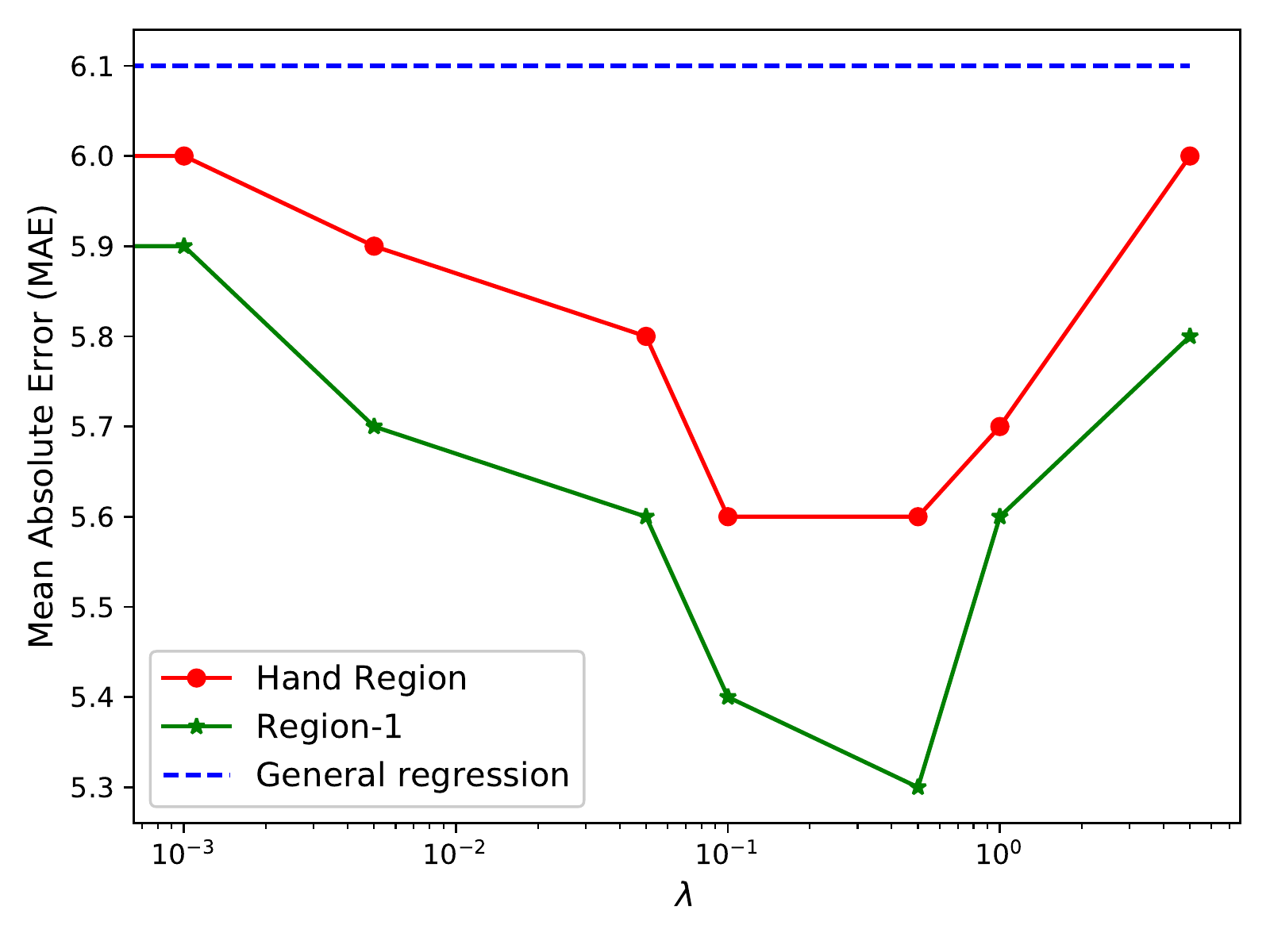}
    \caption{Parameter sensitivity w.r.t. $\lambda$}
    \label{fig52}
  \end{subfigure}
\caption{Parameter sensitivity analysis. Fig (a) shows the MAD w.r.t. threshold $\tau$ when using the single Hand region and Region-1 for BAA. Fig (b) shows the MAD w.r.t. $\lambda$, when using a single Hand region and Region-1 for BAA.}
\label{fig5}
\end{figure}

\noindent\textbf{Aggregating Local Regions for BAA} To evaluate the effectiveness of our proposed discriminative local region based BAA, we perform BAA by utilizing a single local region (Hand region, Region-1 and Region-2) and aggregate different local regions for BAA. In order to demonstrate the contribution of joint age distribution learning and expectation regression, we conduct experiments with both (a) general regression with $\ell_1$ loss and (b) age expectation regression with our proposed loss in Eq. \ref{eq11}. The performance comparison between our proposal and several state-of-the-art deep learning methods \cite{iglovikov2018paediatric,spampinato2017deep,cicero2017hand,escobar2019hand} are shown in Table \ref{tab5}. We observe that: (1) Existing state-of-the-art methods have achieved promising results, however, these methods rely on providing extra annotations, such as hand masks, bounding boxes (denoted as Bbox) or key points to train a segmentation or detection model before bone age regression \cite{iglovikov2018paediatric,escobar2019hand}, or rely on averaging multiple model results to achieve better performance \cite{cicero2017hand,iglovikov2018paediatric}. (2) Our method locates the Hand region, Region-1 and Region-2 only with the image level labels, and achieves very promising results only with the Hand region (H) or Region-1 (R1). (3) Performing BAA based on any local region achieves better performance than utilizing the Original image (O), which shows the effectiveness of our attention guided discriminative region localization and also demonstrates the advantages of using the high-resolution local patches for BAA. Note that only using the Region-2 for BAA performs worse than only using Region-1 or Hand region. This is likely because Region-2 is not as discriminative as Region-1 and the localization accuracy of Region-2 is not as good as Region-1 and Hand regions (according to the quantitative evaluation in Table \ref{tab1}). (4) Aggregating different local regions further improves the final results, with the best results of 4.8 and 4.7 achieved by fusing the "H+R1+R2" and "H+R1+E" ("E" denotes the original image with Region-1 "erased"). Regarding why "H+R1+R2" performs similar as "H+R1+E", we believe the reason is that the localization of Region-2 is not as accurate as Region-1, and using the image with Region-1 being erased will enforce the network to learn from the other parts of the hand image, which provides extra information for BAA. With the general regression model with $\ell_1$ loss, our method does not use any extra annotations, data augmentation, or ensemble strategies, while achieving a performance that is competitive with techniques requiring additional supervision.

\noindent\textbf{Joint Age distribution learning for BAA} Instead of using general regression model which can't make use of the ordinal relationship between neighboring ages, we propose joint age distribution learning and age expectation regression. The performance comparison between general regression and joint age distribution learning and expectation regression is shown in Table \ref{tab5}. As can be seen, our proposed strategy improves the performance consistently. When using the original image as input, our proposal improves the result from $7.3$ to $6.2$. And the best performance we achieved is $4.3$ when combining different local regions for BAA. This result demonstrates the effectiveness of exploring the correlation information between different individual ages by joint age distribution learning. To demonstrate that our method learns the suitable age distribution, we illustrate the learned age distribution of four test samples which are distributed in different age ranges. As can be seen in Fig. 4, the learned distribution is close to the Gaussian distribution when we set a appropriate trade-off parameter $\lambda$, and the age with the highest probability is closer to the ground truth age, which shows the effectiveness of the age distribution learning visually. Note that if we don't use the regularization term (set $\lambda$ to 0), the expectation regression also performs better than the general regression with $\ell_1$ loss, but the learned age distribution will be very sharp, and may easily over-fit to the out-of-distribution samples, while using the regularization term prevents the network from over-confidence and further improves the generalization ability \cite{muller2019does}.

\noindent\textbf{Parameter sensitivity analysis} In order to investigate the sensitivity of the main hyper-parameters involved in our proposal, we conduct empirical parameter sensitivity with respect to the threshold $\tau$ and trade-off parameter $\lambda$. To evaluate the model sensitivity with respect to $\tau$, we perform BAA with a single Hand region and Region-1 which are obtained by different thresholds $\tau\in\{20,30,\cdots,80\}$. As shown in Fig. \ref{fig51}, for the hand region, the best performance was achieved when threshold was set to $\tau\in[20,40]$. For the Region-1, the best performance was achieved when the threshold was set to $\tau\in[40,60]$. As for the sensitivity with respect to $\lambda$, we found that the MAD first decreases and then increases as $\lambda$ increases and shows a bell-shaped curve, which shows that a proper trade-off between age distribution regularization and expectation regression improves the performance. Note that even if we did not regularize the age distribution as a Gaussian distribution, the expectation regression also performs better than the general regression with $\ell_1$ loss. The best performance was obtained when we set $\lambda\in[0.1,1]$

\section{Conclusion}
In order to improve the performance of BAA, existing methods have attempted to exploit local information by providing extra annotations and training a segmentation or detection model before BAA. In this work, we introduce an attention guided method to localize the discriminative local regions with only image-level labels, which is more practical and objective. In particular, we accurately localize the hand region, the carpal bones (Region-1), and the metacarpal bones (Region-2). The results showed that using the localized discriminative region for BAA performs better than using the original image, and it is also suggested that we can utilize the high-resolution carpal bone region alone instead of the full hand region for BAA. We also propose to learn the age distribution and age expectation jointly, which makes use of the ordinal relationship among different individual ages and further improves the performance. To the end, we achieve a similar result on the RSNA bone age dataset compared to those state-of-the-art that need to provide extra annotations. In future work, we will try to integrate the local region localization phase and age expectation regression phase into a unified end-to-end learning framework.

\bibliographystyle{IEEEtran}
\bibliography{IEEEabrv,mylib}
\end{document}